\documentclass{article} 
\usepackage{iclr2017_conference,times}
\iclrfinalcopy
\usepackage{url}

\usepackage{graphicx}
\usepackage{caption,subcaption}
\usepackage{placeins}
\usepackage{amsmath,amssymb}
\usepackage{float}

\usepackage{booktabs}
\usepackage{multirow}

\usepackage{algpseudocode}
\usepackage{algorithm}
\usepackage{listings,xcolor}

\usepackage{tikz}
\usepackage{placeins}
\usetikzlibrary{fit,shapes}

\title{Learning to superoptimize programs}

\author{Rudy Bunel$^{1}$, Alban Desmaison$^{1}$, M. Pawan Kumar$^{1,2}$ \& Philip H.S. Torr$^{1}$ \\
  $^{1}$Department of Engineering Science - University of Oxford \\
  $^{2}$Alan Turing Institute\\
  Oxford, UK\\
  \texttt{\{rudy,alban,pawan\}@robots.ox.ac.uk}, \texttt{philip.torr@eng.ox.ac.uk}\\
  \And
  Pushmeet Kohli\\
  Microsoft Research\\
  Redmond, WA 98052, USA \\
  \texttt{pkohli@microsoft.com}\\
}

\begin{document}

\maketitle

\begin{abstract}
  Code super-optimization is the task of transforming any given program to a
  more efficient version while preserving its input-output behaviour. In some
  sense, it is similar to the paraphrase problem from natural language
  processing where the intention is to change the syntax of an utterance without
  changing its semantics. Code-optimization has been the subject of years of
  research that has resulted in the development of rule-based transformation
  strategies that are used by compilers. More recently, however, a class of
  stochastic search based methods have been shown to outperform these
  strategies. This approach involves repeated sampling of modifications to the
  program from a proposal distribution, which are accepted or rejected based on
  whether they preserve correctness and the improvement they achieve. These
  methods, however, neither learn from past behaviour nor do they try to
  leverage the semantics of the program under consideration. Motivated by this
  observation, we present a novel learning based approach for code
  super-optimization. Intuitively, our method works by learning the proposal
  distribution using unbiased estimators of the gradient of the expected
  improvement. Experiments on benchmarks comprising of automatically generated
  as well as existing (``Hacker's Delight'') programs show that the proposed
  method is able to significantly outperform state of the art approaches for
  code super-optimization.
\end{abstract}

\section{Introduction}
Considering the importance of computing to human society, it is not surprising
that a very large body of research has gone into the study of the syntax and
semantics of programs and programming languages. Code super-optimization is an
extremely important problem in this context. Given a program or a snippet of
source-code, super-optimization is the task of transforming it to a version that
has the same input-output behaviour but can be executed on a target compute
architecture more efficiently. Superoptimization provides a natural benchmark
for evaluating representations of programs. As a task, it requires the
decoupling of the semantics of the program from its superfluous properties, the
exact implementation. In some sense, it is the natural analogue of the
paraphrase problem in natural language processing where we want to change syntax
without changing semantics.

Decades of research has been done on the problem of code optimization
resulting in the development of sophisticated rule-based transformation strategies
that are used in compilers to allow them to perform code optimization.
While modern compilers implement a large set of rewrite rules and are able to
achieve impressive speed-ups, they fail to offer any guarantee of optimality, thus
leaving room for further improvement.  An alternative approach is to search over
the space of all possible programs that are equivalent to the compiler output,
and select the one that is the most efficient. If the search is carried out in a
brute-force manner, we are guaranteed to achieve super-optimization. However,
this approach quickly becomes computationally infeasible as the number of
instructions and the length of the program grows.

In order to efficiently perform super-optimization, recent approaches have
started to use a stochastic search procedure, inspired by Markov Chain Monte
Carlo (MCMC) sampling~\citep{schkufza2013stochastic}. Briefly, the search starts
at an initial program, such as the compiler output. It iteratively suggests
modifications to the program, where the probability of a modification is encoded
in a proposal distribution. The modification is either accepted or rejected with
a probability that is dependent on the improvement achieved. Under certain
conditions on the proposal distribution, the above procedure can be shown, in
the limit, to sample from a distribution over programs, where the probability of
a program is related to its quality. In other words, the more efficient a
program, the more times it is encountered, thereby enabling super-optimization.
Using this approach, high-quality implementations of real programs such as the
Montgomery multiplication kernel from the OpenSSL library were discovered. These
implementations outperformed the output of the \texttt{gcc} compiler and even
expert-handwritten assembly code.

One of the main factors that governs the efficiency of the above stochastic
search is the choice of the proposal distribution. Surprisingly, the state of
the art method, Stoke~\citep{schkufza2013stochastic}, employs a proposal
distribution that is neither learnt from past behaviour nor does it depend on the
syntax or semantics of the program under consideration. We argue that this
choice fails to fully exploit the power of stochastic search. For example,
consider the case where we are interested in performing bitwise operations, as
indicated by the compiler output. In this case, it is more likely that the
optimal program will contain bitshifts than floating point opcodes. Yet, Stoke
will assign an equal probability of use to both types of opcodes.

In order to alleviate the aforementioned deficiency of Stoke, we build a
reinforcement learning framework to estimate the proposal distribution for
optimizing the source code under consideration. The score of the distribution is
measured as the expected quality of the program obtained via stochastic search.
Using training data, which consists of a set of input programs, the parameters
are learnt via the REINFORCE algorithm~\citep{williams1992simple}. We
demonstrate the efficacy of our approach on two datasets. The first is composed
of programs from ``Hacker's Delight''~\citep{warren2002hacker}. Due to the
limited diversity of the training samples, we show that it is possible to learn
a prior distribution (unconditioned on the input program) that outperforms the
state of the art. The second dataset contains automatically generated programs
that introduce diversity in the training samples. We show that, in this more
challenging setting, we can learn a conditional distribution given the initial
program that significantly outperforms Stoke.

\section{Related Works}

\paragraph{Super-optimization}
The earliest approaches for super-optimization relied on brute-force search. By
sequentially enumerating all programs in increasing length
orders~\citep{granlund1992eliminating,massalin1987superoptimizer}, the shortest
program meeting the specification is guaranteed to be found. As expected, this
approach scales poorly to longer programs or to large instruction sets.
The longest reported synthesized program was 12 instructions long,
on a restricted instruction set~\citep{massalin1987superoptimizer}.

Trading off completeness for efficiency, stochastic
methods~\citep{schkufza2013stochastic} reduced the number of programs to test by
guiding the exploration of the space, using the observed quality of programs
encountered as hints. In order to improve the size of solvable instances,
\citet{phothilimthana2016scaling} combined stochastic optimizers with smart
enumerative solvers. However, the reliance of stochastic methods on a generic
unspecific exploratory policy made the optimization blind to the problem at
hand. We propose to tackle this problem by learning the proposal distribution.

\paragraph{Neural Computing}
Similar work was done in the restricted case of finding efficient implementation
of computation of value of degree $k$ polynomials~\citep{efficient-math}. Programs were
generated from a grammar, using a learnt policy to prioritise exploration. This
particular approach of guided search looks promising to us, and is in spirit
similar to our proposal, although applied on a very restricted case.

Another approach to guide the exploration of the space of programs was to make
use of the gradients of differentiable relaxation of programs. \citet{anc}
attempted this by simulating program execution using Recurrent Neural Networks.
However, this provided no guarantee that the network parameters were going to
correspond to real programs. Additionally, this method only had the possibility
of performing local, greedy moves, limiting the scope of possible
transformations. On the contrary, our proposed approach operates directly on
actual programs and is capable of accepting short-term detrimental moves.

\paragraph{Learning to optimize}
Outside of program optimization, applying learning algorithms to improve
optimization procedures, either in terms of results achieved or runtime, is a
well studied subject. \citet{doppa2014hc} proposed imitation learning based
methods to deal with structured output spaces, in a ``Learning to search''
framework. While this is similar in spirit to stochastic search, our setting
differs in the crucial aspect of having a valid cost function instead of
searching for one.

More relevant is the recent literature on learning to optimize.
\citet{li2016learning} and \citet{andrychowicz2016learninggd} learn how to
improve on first-order gradient descent algorithms, making use of neural
networks. Our work is similar, as we aim to improve the optimization process.
However, as opposed to the gradient descent that they learn on a continuous
unconstrained space, our initial algorithm is an MCMC sampler on a discrete
domain.

Similarly, training a proposal distribution parameterized by a Neural Network
was also proposed by \citet{paige2016inference} to accelerate inference in
graphical models. Similar approaches were successfully employed in computer
vision problems where data driven proposals allowed to make inference
feasible~\citep{jampani2015informed,kulkarni2015picture,datadrivenmcmc}. Other
approaches to speeding up MCMC inference include the work of
\citet{salimans2015markov}, combining it with Variational inference.

\newcommand{\mT}{\mathcal{T}}
\newcommand{\mR}{\mathcal{R}}
\newcommand{\cost}[2]{\textnormal{cost}\left(\mathcal{#1}_{#2},\mT\right)}
\newcommand{\move}{\mR \rightarrow \mR^{\prime}}
\section{Learning Stochastic Super-optimization}

\subsection{Stochastic search as a program optimization procedure}
Stoke~\citep{schkufza2013stochastic} performs black-box optimization of a cost function on the space of
programs, represented as a series of instructions. Each instruction is composed
of an opcode, specifying what to execute, and some operands, specifying the
corresponding registers. Each given input program $\mT$ defines a cost function. For a candidate
program $\mR$ called \emph{rewrite}, the goal is to optimize the following cost
function:
\begin{equation}
  \label{eq:cost-fun}
  \cost{R}{} = \omega_e \times \textnormal{eq}(\mR, \mT) + \omega_p \times \textnormal{perf}(\mR)
\end{equation}
The term $\textnormal{eq}(\mathcal{R};\mathcal{T})$ measures how well the
outputs of the rewrite match the outputs of the reference program. This can be
obtained either exactly by running a symbolic validator or approximately by
running test cases. The term $\textnormal{perf}(\mathcal{R})$ is a measure of
the efficiency of the program. In this paper, we consider runtime to be the
measure of this efficiency. It can be approximated by the sum of the latency of
all the instructions in the program. Alternatively, runtime of the program on some
test cases can be used.

To find the optimum of this cost function, Stoke runs an MCMC sampler using the
Metropolis~\citep{metropolis1953equation} algorithm. This allows us to sample from the probability
distribution induced by the cost function:
\begin{equation}
  \label{eq:proba-target}
  p(\mR; \mathcal{T}) = \frac{1}{Z} \exp( - \cost{R}{}) ).
\end{equation}
The sampling is done by proposing random moves from a different proposal distribution:
\begin{equation}
  \label{eq:proposal_distribution}
  \mR^{\prime} \sim q(\  \cdot \ | \mR).
\end{equation}
The cost of the new modified program is evaluated and an acceptance criterion is
computed.
This acceptance criterion
\begin{equation}
  \label{eq:sym-acceptance}
  \alpha( \mR, \mT) = \min \left(1 , \frac{p(\mR^{\prime}; \mT)}{p(\mR; \mT)}\right),
\end{equation}
is then used as the parameter of a Bernoulli distribution from which an
accept/reject decision is sampled. If the move is accepted, the state of the
optimizer is updated to $\mR^{\prime}$. Otherwise, it remains in
$\mR$.

While the above procedure is only guaranteed to sample from the distribution
$p(\ \cdot\ ; \mathcal{T})$ in the limit if the proposal distribution $q$ is
symmetric ($q(\mR^{\prime}|\mR) = q(\mR|\mR^{\prime})$ for all $\mR,
\mR^{\prime}$), it still allows us to perform efficient hill-climbing for
non-symmetric proposal distributions. Moves leading to an improvement are always
going to be accepted, while detrimental moves can still be accepted in order to
avoid getting stuck in local minima.

\subsection{Learning to search}
We now describe our approach to improve stochastic search by learning the
proposal distribution. We begin our description by defining the learning
objective (section~\ref{sssec:objective}), followed by a parameterization of the
proposal distribution (section~\ref{sssec:params}), and finally the
reinforcement learning framework to estimate the parameters of the proposal
distribution (section~\ref{sssec:estimate}).

\subsubsection{Objective function}
\label{sssec:objective}
Our goal is to optimize the cost function defined in
equation~(\ref{eq:cost-fun}). Given a fixed computational budget of $T$
iterations to perform program super-optimization, we want to make moves that
lead us to the lowest possible cost. As different programs have different
runtimes and therefore different associated costs, we need to perform
normalization. As normalized loss function, we use the ratio between the best
rewrite found and the cost of the initial unoptimized program $\mR_0$. Formally,
the loss for a set of rewrites $\{\mR_t\}_{t=0..T}$ is defined as follows:
\begin{equation}
  \label{eq:reward_def}
  r(\{\mR_t\}_{t=0..T}) = \left ( \frac{\min_{t=0..T}\cost{R}{t}}{\cost{R}{0}} \right ).
\end{equation}

Recall that our goal is to learn a proposal distribution. Given that our
optimization procedure is stochastic, we will need to consider the expected cost
as our loss. This expected loss is a function of the parameters $\theta$ of our
parametric proposal distribution $q_\theta$:
\begin{equation}
  \label{eq:objective}
  \mathcal{L}(\theta) = \mathbb{E}_{\{\mR_t\} \sim q_{\theta}}\left[ r(\{\mR_t\}_{t=0..T}) \right].
\end{equation}

\subsubsection{Parameterization of the Move Proposal Distribution}
\label{sssec:params}
The proposal distribution (\ref{eq:proposal_distribution}) originally used in Stoke
\citep{schkufza2013stochastic} takes the form of a hierarchical model. The type
of the move is initially sampled from a probability distribution. Additional
samples are drawn to specify, for example, the affected location in the programs
,the new operands or opcode to use. Which of these probability distribution get
sampled depends on the type of move that was first sampled. The detailed
structure of the proposal probability distribution can be found in
Appendix~\ref{sec:proba_struct}.

Stoke uses uniform distributions for each of the elementary probability
distributions the model samples from. This corresponds to a specific
instantiation of the general stochastic search paradigm. In this work, we
propose to learn those probability distributions so as to maximize the
probability of reaching the best programs. The rest of the optimization scheme
remains similar to the one of \citet{schkufza2013stochastic}.

Our chosen parameterization of $q$ is to keep the hierarchical structure of the
original work of~\citet{schkufza2013stochastic}, as detailed in
Appendix~\ref{sec:proba_struct}, and parameterize all the elementary probability
distributions (over the positions in the programs, the instructions to propose
or the arguments) independently. The set $\theta$ of parameters for $q_\theta$
will thus contain a set of parameters for each elementary probability
distributions. A fixed proposal distribution is kept through the optimization of
a given program, so the proposal distribution needs to be evaluated only once,
at the beginning of the optimization and not at every iteration of MCMC.

\begin{figure}[h]
  \centering
  \begin{tikzpicture}

  \node[draw](feat) at (0,-0.5) {Feature of original program};
  \node[draw, ultra thick](proposal) at (0, 1) {Proposal Distribution};
  \draw[->, draw=red] ([xshift=-.5cm]feat.north) to node[midway,left] {\color{red}Neural Network (a)} ([xshift=-.5cm]proposal.south);
  \draw[->, draw=green] ([xshift=.5cm]proposal.south) to node[midway,right] {\color{green}BackPropagation} ([xshift=.5cm]feat.north);

  \node[draw, ellipse](move) at (0,3) {Move};
   \draw[dashed,->, draw=red] ([xshift=-.5cm]proposal.north) to node[ midway,left](catsamp) {\color{red}Categorical Sample (b)} ([xshift=-.5cm]move.south);
  \draw[dashed, ->, draw=green] ([xshift=.5cm]move.south) to node[midway,right](reinf) {\color{green}REINFORCE} ([xshift=.5cm]proposal.north);

  \node[draw](prog) at (4,4) {Program};

  \node[draw](candidate) at (0,4.5) {Candidate Rewrite};
  \draw[->] (move) to node[midway,left]{} (candidate);
  \draw[->] (prog) to (candidate);

  \node[draw](newscore) at (0,6) {Candidate score};
  \draw[->](candidate) to node[midway,left]{(c)} (newscore);

  \node[draw](currscore) at (2.5, 6) {Score};
  \draw[->] (prog) to (currscore);

  \node[draw](acceptance) at (2.5, 7.5) {Acceptance criterion};
  \draw[->] (newscore) to node[midway,left]{(d)} ([xshift=-1.5cm]acceptance.south);
  \draw[->] (currscore) to node[midway,left]{(d)} (acceptance);

  \node[draw, ellipse](rewrite) at (2.5, 9) {New rewrite};
  \draw[dashed, ->] (acceptance.north) to node[midway,left] {Bernoulli (e)} (rewrite.south);

  \draw[->] (rewrite.east) to[out=0,in=90] node[very near end, right]{(g)} ([xshift=4cm]candidate.east)
                           to[out=-90, in=0] (prog.east);

  \node[draw,fill=yellow](reward) at (2.5, 10.5) {Cost};
  \draw[->] (rewrite) to node[near start,left](cost_label){(f)} (reward);

  \node[draw, dotted, fit=(move) (prog) (rewrite) (candidate) (newscore) (acceptance) (cost_label) (reinf) (catsamp)]{};
\end{tikzpicture}
  \caption{\em{Stochastic computation graph of the Metropolis algorithm used
    for program super-optimization.  Round nodes are stochastic nodes and square
    ones are deterministic. Red arrows corresponds to computation done in
  the forward pass that needs to be learned while green arrows correspond to the
backward pass. Full arrows represent deterministic computation and dashed arrows
represent stochastic ones. The different steps of the forward pass are:\\
(a) Based on features of the reference program, the proposal distribution $q$
  is computed.\\
(b) A random move is sampled from the proposal distribution. \\
(c) The score of the proposed rewrite is experimentally measured. \\
(d) The acceptance criterion (\ref{eq:sym-acceptance}) for the move is computed.\\
(e) The move is accepted with a probability equal to the acceptance criterion.\\
(f) The cost is observed, corresponding to the best program obtained during
the search.\\
(g) Moves b to f are repeated $T$ times.\\
}}
  \label{fig:m-stocha-compg}
\end{figure}
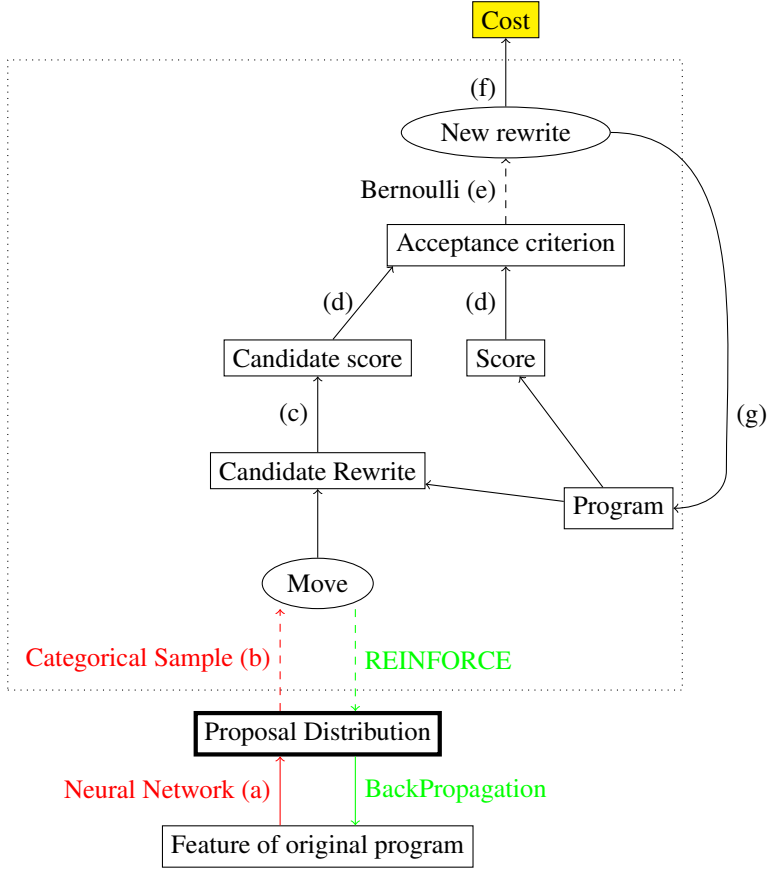

The stochastic computation graph corresponding to a run of the
Metropolis algorithm is given in Figure~\ref{fig:m-stocha-compg}. We have
assumed the operation of evaluating the cost of a program to be a deterministic
function, as we will not model the randomness of measuring performance.

\subsubsection{Learning the Proposal Distribution}
\label{sssec:estimate}
In order to learn the proposal distribution, we will use stochastic gradient descent on
our loss function~(\ref{eq:objective}). We obtain the first order derivatives with
regards to our proposal distribution parameters using the
REINFORCE~\citep{williams1992simple} estimator, also known as the likelihood
ratio estimator~\citep{glynn1990likelihood} or the score function estimator~\citep{fu2006575}.
This estimator relies on a rewriting of the gradient of the expectation. For an
expectation with regards to a probability distribution $x \sim f_\theta$, the
REINFORCE estimator is:
\begin{equation}
  \label{eq:reinforce}
  \nabla_\theta \sum_{x} f(x; \theta) r(x) = \sum_{x} r(x) \nabla_\theta f(x; \theta) = \sum_{x} f(x; \theta) r(x) \nabla_\theta \log(f(x; \theta)),
\end{equation}
and provides an unbiased estimate of the gradient.

A helpful way to derive the gradients is to consider the execution traces of the
search procedure under the formalism of stochastic computation
graphs~\citep{schulman2015gradient}. We introduce one ``cost node'' in the
computation graphs at the end of each iteration of the sampler. The associated
cost corresponds to the normalized difference between the best rewrite
so far and the current rewrite after this step:
\begin{equation}
 \label{eq:cost_node}
 c_{t} = \min \left( 0,  \left ( \frac{\cost{R}{t}- \min_{i=0..t-1}\cost{R}{i} }{\cost{R}{0}} \right ) \right).
\end{equation}

The sum of all the cost nodes corresponds to the sum of all the improvements
made when a new lowest cost was achieved. It can be shown that up to a constant
term, this is equivalent to our objective function~(\ref{eq:reward_def}). As
opposed to considering only a final cost node at the end of the $T$ iterations,
this has the advantage that moves which were not responsible for the
improvements would not get assigned any credit.

For each round of MCMC, the gradient with regards to the proposal
distribution is computed using the REINFORCE estimator which is equal to
\begin{equation}
  \label{eq:proposal_grad}
  \widehat{\nabla_{\theta, i }} \mathcal{L}(\theta)= (\nabla_\theta \log q_\theta(\mR_i | \mR_{i-1})) \sum_{t>i} c_{t}.
\end{equation}
As our proposal distribution remains fixed for the duration of a program
optimization, these gradients needs to be summed over all the iterations to
obtain the total contribution to the proposal distribution. Once this gradient
is estimated, it becomes possible to run standard back-propagation
with regards to the features on which the proposal distribution is based on, so as to
learn the appropriate feature representation.

\section{Experiments}
\subsection{Setup}
\paragraph{Implementation}
Our system is built on top of the Stoke super-optimizer from
\citet{schkufza2013stochastic}. We instrumented the implementation of the
Metropolis algorithm to allow sampling from parameterized proposal distributions
instead of the uniform distributions previously used. Because the proposal distribution is
only evaluated once per program optimisation, the impact on the optimization
throughput is low, as indicated in Table~\ref{tab:run-throughput}.

Our implementation also keeps track of the traces through the stochastic graph.
Using the traces generated during the optimization, we can compute the estimator
of our gradients, implemented using the Torch framework~\citep{torch}.

\paragraph{Datasets}
We validate the feasibility of our learning approach on two experiments. The
first is based on the Hacker's delight~\citep{warren2002hacker} corpus, a
collection of twenty five bit-manipulation programs, used as benchmark in program
synthesis~\citep{gulwani11loopfree,jha2010oracle,schkufza2013stochastic}. Those
are short programs, all performing similar types of tasks. Some examples
include identifying whether an integer is a power of two from its binary
representation, counting the number of bits turned on in a register or computing
the maximum of two integers. An exhaustive description of the tasks is given in
Appendix~\ref{sec:hd_task}. Our second corpus of programs is automatically
generated and is more diverse.

\paragraph{Models}
The models we are learning are a set of simple elementary probabilities for
the categorical distribution over the instructions and over the type of moves to
perform. We learn the parameters of each separate distribution jointly, using a
Softmax transformation to enforce that they are proper probability distributions.
For the types of move where opcodes are chosen from a specific subset, the
probabilities of each instruction are appropriately renormalized.
We learn two different type of models and compare them with the baseline of
uniform proposal distributions equivalent to Stoke.

Our first model, henceforth denoted the bias, is not conditioned on any property
of the programs to optimize. By learning this simple proposal distribution, it
is only possible to capture a bias in the dataset. This can be
understood as an optimal proposal distribution that Stoke should default to.

The second model is a Multi Layer Perceptron (MLP), conditioned on the input program
to optimize. For each input program, we generate a Bag-of-Words representation
based on the opcodes of the program. This is embedded through a three hidden
layer MLP with ReLU activation unit. The proposal distribution over the
instructions and over the type of moves are each the result of passing the
outputs of this embedding through a linear transformation, followed by a
SoftMax.

The optimization is performed by stochastic gradient descent, using the
Adam~\citep{kingma2014adam} optimizer. For each estimate of the gradient, we
draw 100 samples for our estimator. The values of the hyperparameters used are
given in Appendix \ref{sec:hyperparams}. The number of parameters of each model
is given in Table~\ref{tab:model_params}.

\subsection{Existing Programs}
In order to have a larger corpus than the twenty-five programs initially
present in ``Hacker's Delight'', we generate various starting points for each optimization. This is
accomplished by running Stoke with a cost function where $\omega_p = 0$ in
(\ref{eq:cost-fun}), and keeping only the correct programs. Duplicate programs
are filtered out. This allows us to create a larger dataset from which to learn.
Examples of these programs at different level of optimization can be found in
Appendix~\ref{sec:hd_opt}.

We divide this augmented Hacker's Delight dataset into two sets. All the
programs corresponding to even-numbered tasks are assigned to the first set,
which we use for training. The programs corresponding to odd-numbered tasks are
kept for separate evaluation, so as to evaluate the generalisation of our
learnt proposal distribution.

\begin{figure}[h]
  \centering
  \hfill%
  \begin{subfigure}{0.30\linewidth}
    \includegraphics[width=\linewidth]{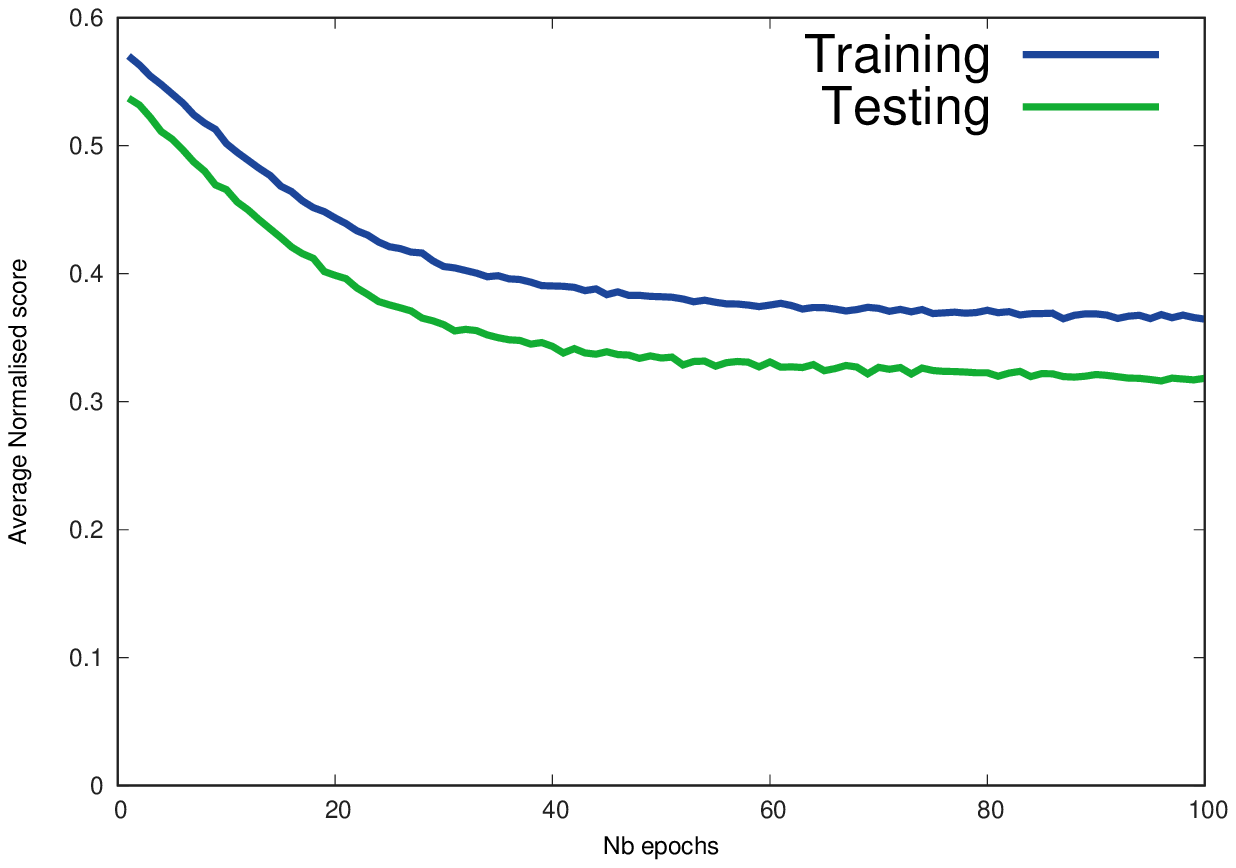}%
    \caption{\label{fig:hd_train_bias}Bias}
  \end{subfigure}%
  \hfill%
  \begin{subfigure}{0.30\linewidth}
    \includegraphics[width=\linewidth]{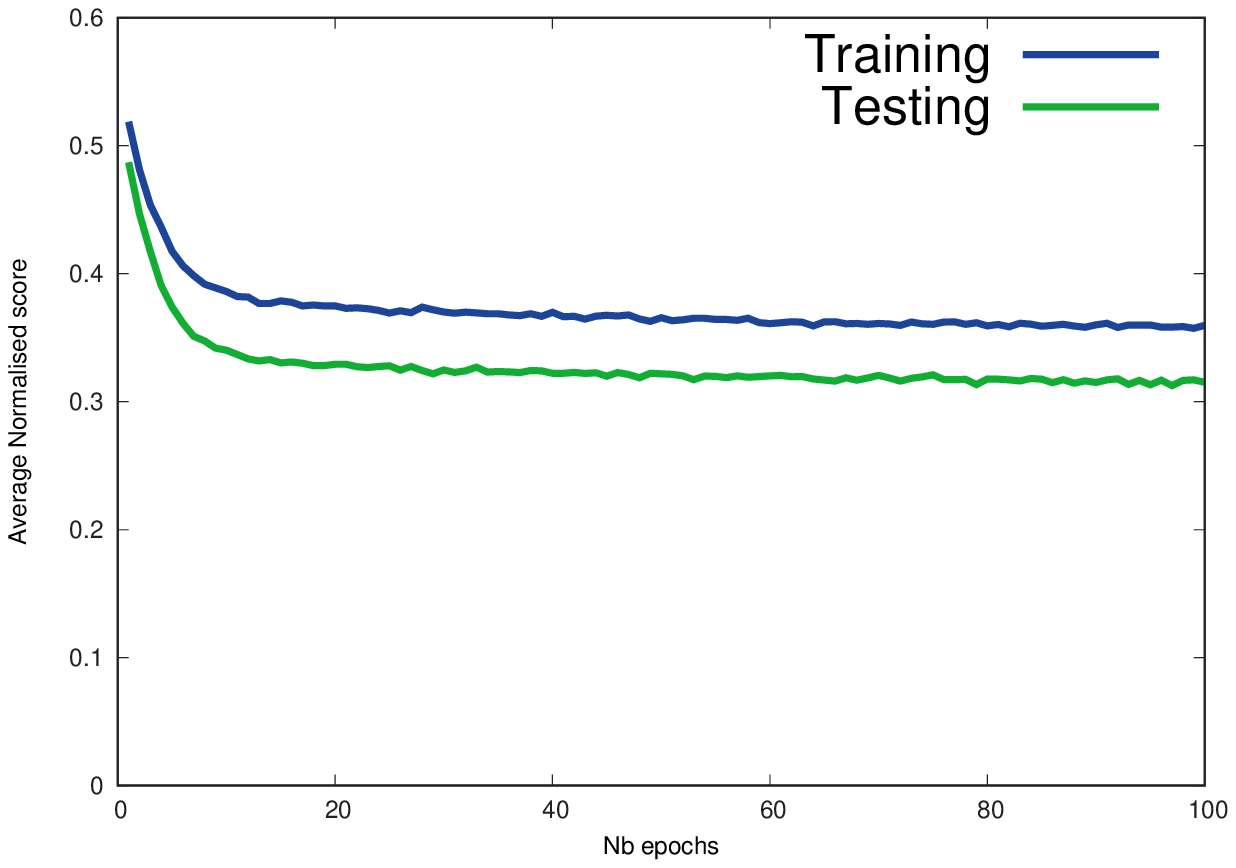}
    \caption{\label{fig:hd_train_mlp}Multi-layer Perceptron}
  \end{subfigure}%
  \hspace*{\fill}
  \caption{\label{fig:hd_res}Proposal distribution training. All models learn
    to improve the performance of the stochastic optimization. Because the tasks
    are different between the training and testing dataset, the values between
    datasets can't directly be compared as some tasks have more opportunity for
    optimization. It can however be noted that improvements on the training
    dataset generalise to the unseen tasks.}
\end{figure}

The optimization process is visible in Figure~\ref{fig:hd_res}, which shows a
clear decrease of the training loss and testing loss for both models. While
simply using stochastic super-optimization allows to discover programs 40\% more
efficient on average, using a tuned proposal distribution yield even larger
improvements, bringing the improvements up to 60\%, as can be seen in
Table\ref{tab:hd_comp}. Due to the similarity between the different tasks,
conditioning on the program features does not bring any significant
improvements.

\begin{table}
  \parbox[t]{.3\linewidth}{
    \renewcommand{\arraystretch}{1.4}
    \begin{tabular}[t]{|c|c|}
      \hline
      {\bf Model} & {\bf \# of parameters}\\
      \hline
      Uniform & $0$ \\
      Bias & $2912$\\
      MLP & $1.4 \times 10^6 $\\ 
      \hline
    \end{tabular}
    \caption{\label{tab:model_params}Size of the different models compared.\\
      Uniform corresponds to Stoke \cite{schkufza2013stochastic}.}
    }
    \hfill
    \parbox[t]{.65\linewidth}{
      \centering
      \renewcommand{\arraystretch}{1.2}
      \begin{tabular}[t]{c@{\hspace*{6ex}}cc}
        \toprule
      \textbf{Model} & \textbf{Training}  & \textbf{Test} \\
        \midrule
        Uniform & 57.01\% & 53.71\% \\
        Bias & 36.45 \% & 31.82 \% \\
        MLP & \underline{35.96 \%} & \underline{31.51 \%}\\
        \bottomrule
      \end{tabular}
      \caption{\label{tab:hd_comp}Final average relative score on the Hacker's
        Delight benchmark. While all models improve with regards to the initial
        proposal distribution based on uniform sampling, the model conditioning
        on program features reach better performances.}
    }
\end{table}

\begin{figure}[h]
  \begin{subfigure}{0.3\linewidth}
    \captionsetup{justification=centering}
    \centering
    \includegraphics[width=0.8\linewidth]{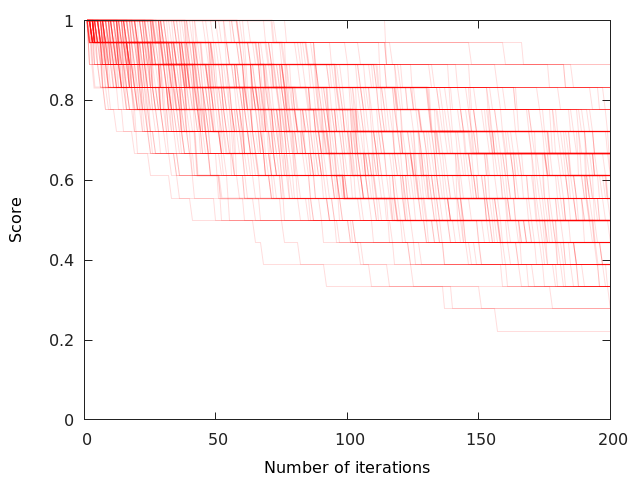}
    \caption{\label{fig:bias_before_train_traces} With Uniform proposal\\ Optimization Traces}
  \end{subfigure}%
  \hfill%
  \begin{subfigure}{0.3\linewidth}
    \centering
    \includegraphics[width=0.8\linewidth]{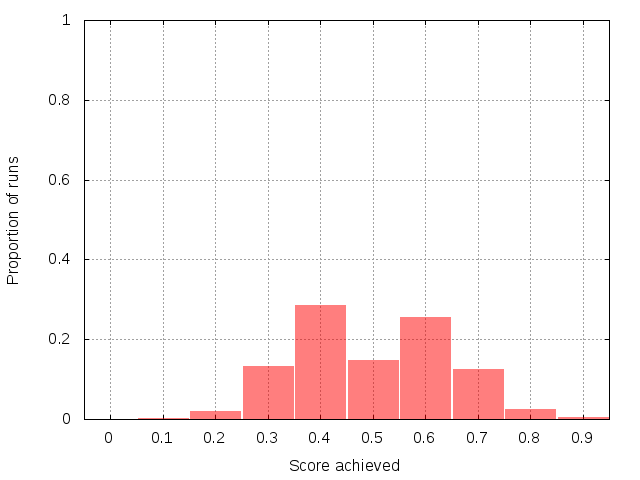}
    \caption{\label{fig:uni_200}Scores after 200 iterations}
  \end{subfigure}
  \hfill%
  \begin{subfigure}{0.3\linewidth}
    \centering
    \includegraphics[width=0.8\linewidth]{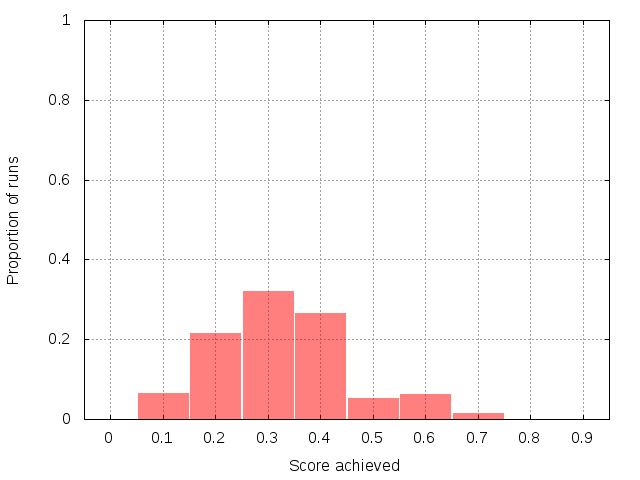}
    \caption{\label{fig:uni_400}Scores after 400 iterations}
  \end{subfigure}

 \begin{subfigure}{0.3\linewidth}
   \captionsetup{justification=centering}
   \centering
   \includegraphics[width=0.8\linewidth]{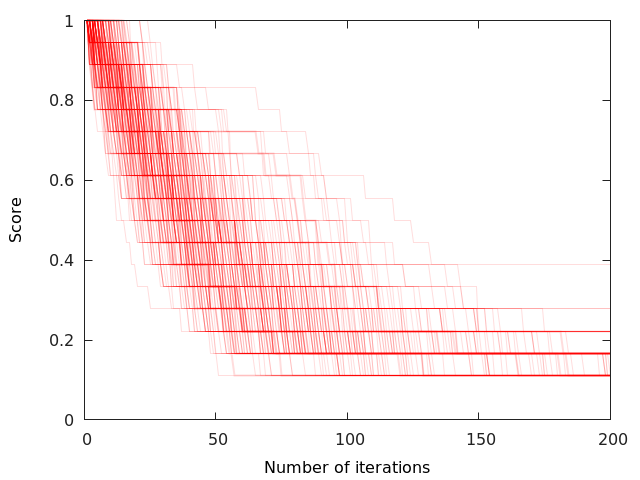}
   \caption{\label{fig:bias_after_train_traces} With Learned Bias\\ Optimization Traces}
 \end{subfigure}%
 \hfill%
 \begin{subfigure}{0.3\linewidth}
   \centering
   \includegraphics[width=0.8\linewidth]{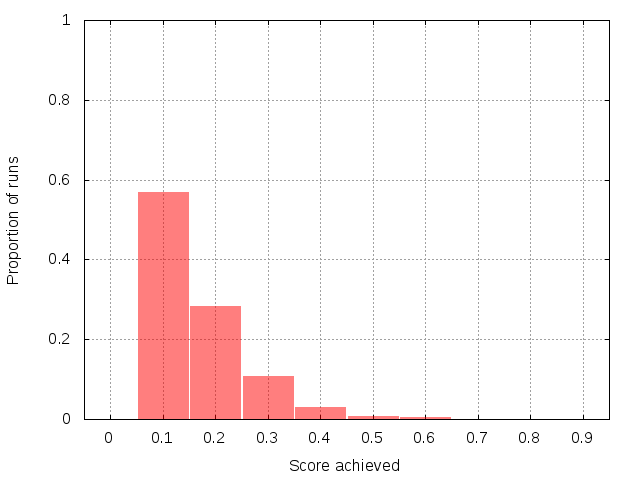}
   \caption{\label{fig:bias_100}Scores after 100 iterations}
 \end{subfigure}
 \hfill%
 \begin{subfigure}{0.3\linewidth}
   \centering
   \includegraphics[width=0.8\linewidth]{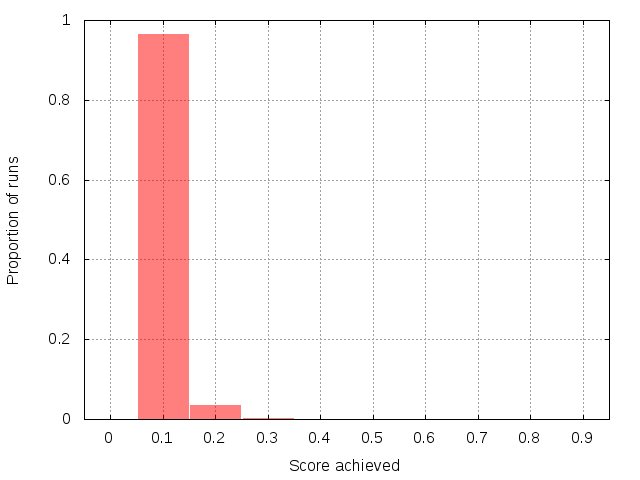}
   \caption{\label{fig:bias_200}Scores after 200 iterations}
 \end{subfigure}

 \caption{\label{fig:hd_traces}Distribution of the improvement achieved when
   optimising a training sample from the Hacker's Delight dataset. The first
   column represent the evolution of the score during the optimization. The
   other columns represent the distribution of scores after a given number of iterations.\\
   (a) to (c) correspond to the uniform proposal distribution,
   (d) to (f) correspond to the learned bias.
 }
\end{figure}

In addition, to clearly demonstrate the practical consequences of our learning,
we present in Figure~\ref{fig:hd_traces} a superposition of score traces,
sampled from the optimization of a program of the test set.
Figure~\ref{fig:bias_before_train_traces} corresponds to our initialisation, an
uniform distribution as was used in the work of \citet{schkufza2013stochastic}.
Figure~\ref{fig:bias_after_train_traces} corresponds to our optimized version.
It can be observed that, while the uniform proposal distribution was
successfully decreasing the cost of the program, our learnt proposal
distribution manages to achieve lower scores in a more robust manner and in less
iterations. Even using only 100 iterations (Figure \ref{fig:bias_100}), the
learned model outperforms the uniform proposal distribution with 400 iterations
(Figure \ref{fig:uni_400}).

\subsection{Automatically Generated Programs}
While the previous experiments shows promising results on a set of programs of
interest, the limited diversity of programs might have made the task too simple,
as evidenced by the good performance of a blind model. Indeed, despite the data
augmentation, only 25 different tasks were present, all variations of
the same programs task having the same optimum.

To evaluate our performance on a more challenging problem, we automatically synthesize
a larger dataset of programs. Our methods to do so consists in running Stoke
repeatedly with a constant cost function, for a large number of iterations. This
leads to a fully random walk as every proposed programs will have the same
cost, leading to a 50\% chance of acceptance. We generate 600 of these programs,
300 that we use as a training set for the optimizer to learn over and 300 that
we keep as a test set.

\begin{figure}[h!]
  \centering
  \hfill%
  \begin{subfigure}{0.30\linewidth}
    \includegraphics[width=\linewidth]{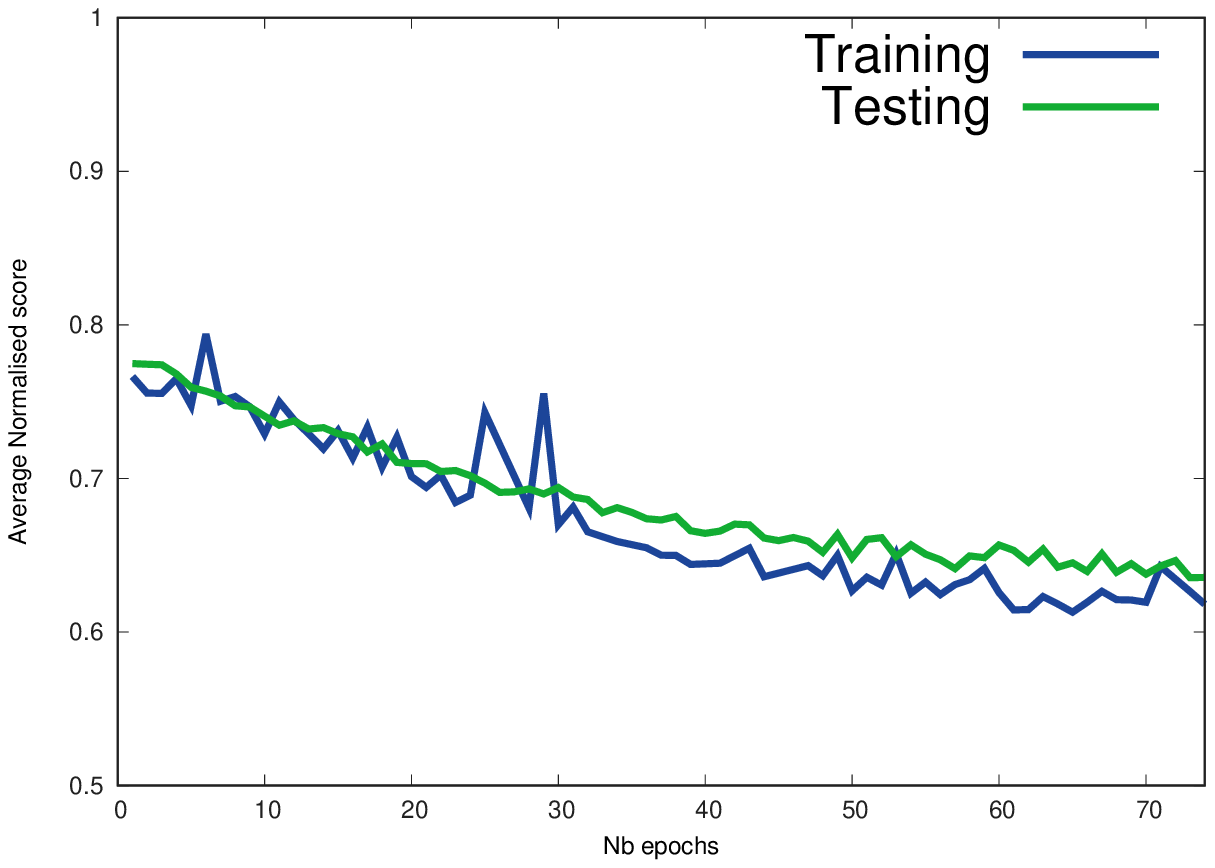}%
    \caption{\label{fig:synth_train_bias}Bias}
  \end{subfigure}%
  \hfill%
  \begin{subfigure}{0.30\linewidth}
    \includegraphics[width=\linewidth]{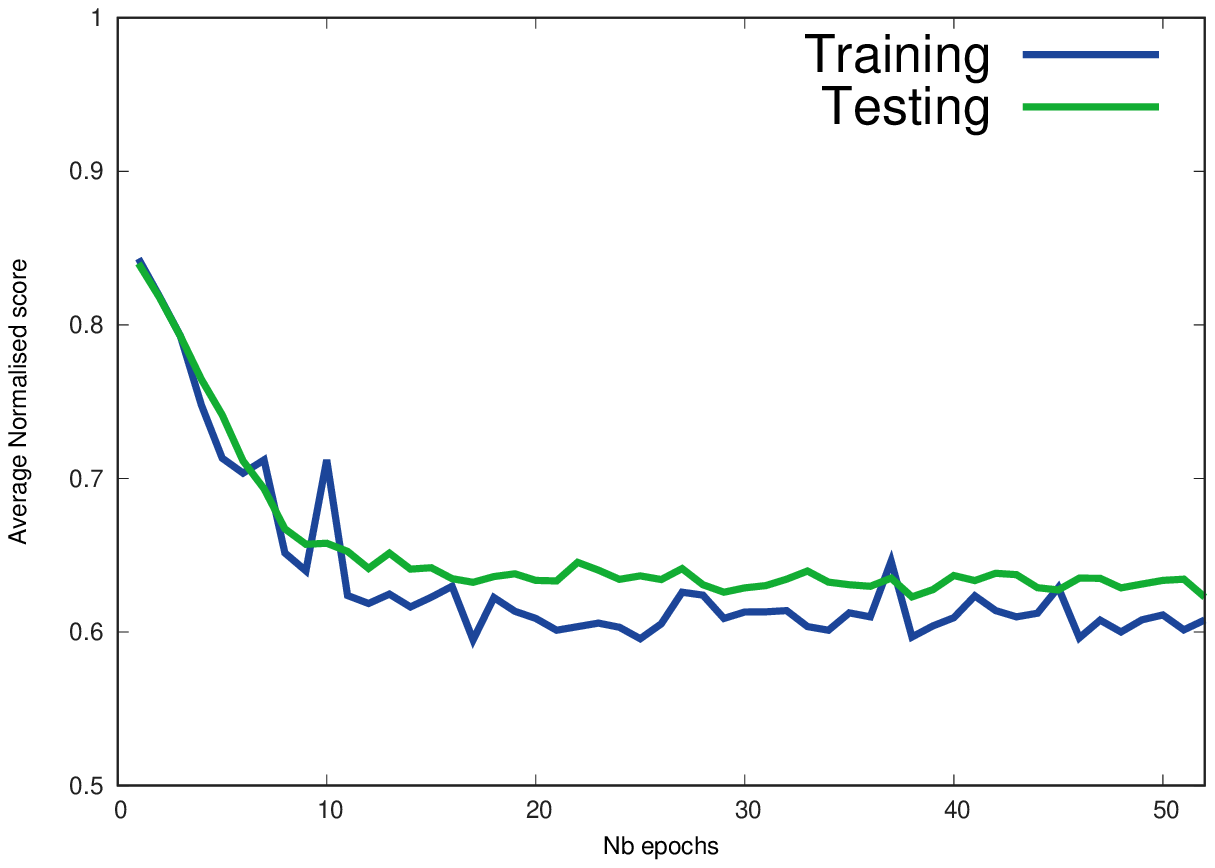}
    \caption{\label{fig:synth_train_mlp}Multi-layer Perceptron}
  \end{subfigure}%
  \hspace*{\fill}
  \caption{\label{fig:synth_res}Training of the proposal distribution on the
    automatically generated benchmark.
  }
\end{figure}

The performance achieved on this more complex dataset is shown in
Figure~\ref{fig:synth_res} and Table~\ref{tab:synth_comp}.

\begin{table}
  \parbox[t]{.4\linewidth}{
    \renewcommand{\arraystretch}{1.2}
    \begin{tabular}[t]{|c|c|}
      \hline
      \begin{tabular}{@{}c@{}}\bf Proposal  \\ \bf distribution\end{tabular} & \begin{tabular}{@{}c@{}}\bf MCMC iterations  \\ \bf throughput\end{tabular} \\
      \hline
      Uniform & 60 000 /second \\
      Categorical & 20 000 /second \\
      \hline
    \end{tabular}
    \caption{\label{tab:run-throughput}Throughput of the proposal distribution
      estimated by timing MCMC for 10000 iterations}
  }
  \hfill
  \parbox[t]{.55\linewidth}{
    \centering
    \renewcommand{\arraystretch}{1.2}
    \begin{tabular}[t]{c@{\hspace*{6ex}}cc}
      \toprule
      \textbf{Model} & \textbf{Training}  & \textbf{Test} \\
    \midrule
      Uniform & 76.63\% & 78.15 \% \\
      Bias & 61.81\% & 63.56\% \\
      MLP & \underline{60.13\%} & \underline{62.27\%}\\
      \bottomrule
    \end{tabular}
    \caption{\label{tab:synth_comp}Final average relative score. The MLP
      conditioning on the features of the program perform better than the
      simple bias. Even the unconditioned bias performs significantly better
      than the Uniform proposal distribution.}
  }
\end{table}

\FloatBarrier
\section{Conclusion}
Within this paper, we have formulated the problem of optimizing the performance
of a stochastic super-optimizer as a Machine Learning problem. We demonstrated
that learning the proposal distribution of a MCMC sampler was feasible and lead
to faster and higher quality improvements. Our approach is not limited to
stochastic superoptimization and could be applied to other stochastic search
problems.

It is interesting to compare our method to the synthesis-style approaches that
have been appearing recently in the Deep Learning
community~\citep{graves2014neural} that aim at learning algorithms directly
using differentiable representations of programs. We find that the stochastic
search-based approach yields a significant advantage compared to those types of
approaches, as the resulting program can be run independently from the Neural
Network that was used to discover them.

Several improvements are possible to the presented methods. In mature domains
such as Computer Vision, the representations of objects of interests have been
widely studied and as a result are successful at capturing the information of
each sample. In the domains of programs, obtaining informative representations
remains a challenge. Our proposed approach ignores part of the structure of the
program, notably temporal, due to the limited amount of existing data. The
synthetic data having no structure, it wouldn't be suitable to learn those
representations from it. Gathering a larger dataset of frequently used
programs so as to measure more accurately the practical performance of those
methods seems the evident next step for the task of program synthesis.

\section{Acknowledgments}
This work was supported by the EPSRC, ERC grant ERC-2012-AdG 321162-HELIOS,
EPSRC grant Seebibyte EP/M013774/1, EPSRC/MURI grant EP/N019474/1 and
Microsoft scolarships.

\clearpage
{\footnotesize
\bibliographystyle{iclr2017_conference}
\bibliography{bibliography}
}
\appendix
\section{Hyperparameters}
\label{sec:hyperparams}

\subsection{Architectures}

The output size of 9 corresponds to the types of move. The output size of 2903
correspond to the number of possible instructions that Stoke can use during a
rewrite. This is smaller that the 3874 that are possible to find in an original
program.
\begin{table}[H]
  \centering
  \begin{tabular}{|l|c|c|}
    \hline
    \multirow{2}{*}{Outputs}
    & Bias (9) & Bias (2903) \\
    & SoftMax & SoftMax\\
    \hline
  \end{tabular}
  \caption{\label{tab:mlp_arch}Architecture of the Bias}
\end{table}

\begin{table}[H]
  \centering
  \begin{tabular}{|l|c|c|}
    \hline
    \multirow{3}{*}{Embedding}
    & \multicolumn{2}{|c|}{Linear (3874 $\rightarrow$ 100) + ReLU} \\
    & \multicolumn{2}{|c|}{Linear (100 $\rightarrow$ 300) + ReLU} \\
    & \multicolumn{2}{|c|}{Linear (300 $\rightarrow$ 300) + ReLU} \\
    \hline
    \multirow{2}{*}{Outputs}
    & Linear (300 $\rightarrow$ 9) & Linear (300 $\rightarrow$ 2903) \\
    & SoftMax & SoftMax\\
    \hline
  \end{tabular}
  \caption{\label{tab:mlp_arch}Architecture of the Multi Layer Perceptron}
\end{table}

\subsection{Training parameters}
All of our models are trained using the Adam~\citep{kingma2014adam} optimizer,
with its default hyper-parameters $\beta_1=0.9$, $\beta_2=0.999$,
$\epsilon=10^{-8}$. We use minibatches of size 32.

The learning rate were tuned by observing the evolution of the loss on the
training datasets for the first iterations. The picked values are given in Table
\ref{tab:lr}. Those learning rates are divided by the size of the minibatches.

\begin{table}[H]
  \centering
  \begin{tabular}{|l|c|c|}
    \hline
    & Hacker's Delight & Synthetic\\
    \hline
    Bias & 1 & 10\\
    MLP & 0.01 & 0.1 \\
    \hline
  \end{tabular}
  \caption{\label{tab:lr}Values of the Learning rate used.}
\end{table}

\section{Structure of the proposal distribution}
\label{sec:proba_struct}
The sampling process of a move is a hierarchy of sampling step. The easiest way
to represent it is as a generative model for the program transformations.
Depending on what type of move is sampled, different series of sampling steps
have to be performed. For a given move, all the probabilities are sampled
independently so the probability of proposing the move is the product of the
probability of picking each of the sampling steps. The generative model is
defined in Figure~\ref{lst:genCode}. It is going to be parameterized by the the
parameters of each specific probability distribution it samples from. The
default Stoke version uses uniform probabilities over all of those elementary
distributions.

\begin{figure}
  \lstdefinelanguage{ppl}
  {
    keywords={sample},
    keywordstyle=\color{blue}\bfseries,
    morekeywords=[2]{def},
    keywordstyle=[2]\color{green}\bfseries,
    morekeywords=[3]{if,return},
    keywordstyle=[3]\bfseries,
    comment=[l]{\%},
    commentstyle=\color{orange}\ttfamily
  }
  \lstset{
    language={ppl},
    basicstyle=\normalfont\ttfamily\footnotesize,
    columns=fixed,
    frame = trbl,
    numbers=left,
    numberstyle=\tiny\ttfamily,
  }
  \newsavebox{\temploclistingbox}
  \begin{lrbox}{\temploclistingbox}%
    \begin{minipage}{\textwidth}%
\begin{lstlisting}[language=ppl]
def proposal(current_program):
    move_type = sample(categorical(all_move_type))
    if move_type == 1: % Add empty Instruction
       pos = sample(categorical(all_positions(current_program)))
       return (ADD_NOP, pos)

    if move_type == 2: % Delete an Instruction
       pos = sample(categorical(all_positions(current_program)))
       return (DELETE, pos)

    if move_type == 3: % Instruction Transform
       pos = sample(categorical(all_positions(current_program)))
       instr = sample(categorical(set_of_all_instructions))
       arity = nb_args(instr)
       for i = 1, arity:
           possible_args = possible_arguments(instr, i)
           % get one of the arguments that can be used as i-th
           % argument for the instruction 'instr'.
           operands[i] = sample(categorical(possible_args))
       return (TRANSFORM, pos, instr, operands)

    if move_type == 4: % Opcode Transform
       pos = sample(categorical(all_positions(current_program)))
       args = arguments_at(current_program, pos)
       instr = sample(categorical(possible_instruction(args)))
       % get an instruction compatible with the arguments
       % that are in the program at line pos.
       return(OPCODE_TRANSFORM, pos, instr)

    if move_type == 5: % Opcode Width Transform
       pos = sample(categorical(all_positions(current_program))
       curr_instr = instruction_at(current_program, pos)
       instr = sample(categorical(same_memonic_instr(curr_instr))
       % get one instruction with the same memonic that the
       % instruction 'curr_instr'.
       return (OPCODE_TRANSFORM, pos, instr)

    if move_type == 6: % Operand transform
       pos = sample(categorical(all_positions(current-program))
       curr_instr = instruction_at(current_program, pos)
       arg_to_mod = sample(categorical(args(curr_instr)))
       possible_args = possible_arguments(curr_instr, arg_to_mod)
       new_operand = sample(categorical(possible_args))
       return (OPERAND_TRANSFORM, pos, arg_to_mod, new_operand)

    if move_type == 7: % Local swap transform
       block_idx = sample(categorical(all_blocks(current_program)))
       possible_pos = pos_in_block(current_program, block_idx)
       pos_1 = sample(categorical(possible_pos))
       pos_2 = sample(categorical(possible_pos))
       return (SWAP, pos_1, pos_2)

    if move_type == 8: % Global swap transform
       pos_1 = sample(categorical(all_positions(current_program)))
       pos_2 = sample(categorical(all_positions(current_program)))
       return (SWAP, pos_1, pos_2)

    if move_type == 9: % Rotate transform
       pos_1 = sample(categorical(all_positions(current_program)))
       pos_2 = sample(categorical(all_positions(current_program)))
       return (ROTATE, pos_1, pos_2)
\end{lstlisting}
    \end{minipage}%
  \end{lrbox}%
  \usebox{\temploclistingbox}
  \caption{\label{lst:genCode}Generative Model of a Transformation.\\
  }
\end{figure}

\FloatBarrier
\section{Hacker's Delight Tasks}
\label{sec:hd_task}
The 25 tasks of the Hacker's delight~\cite{warren2002hacker} datasets are the following:

\begin{enumerate}
\item Turn off the right-most one bit
\item Test whether an unsigned integer is of the form $2^(n-1)$
\item Isolate the right-most one bit
\item Form a mask that identifies right-most one bit and trailing zeros
\item Right propagate right-most one bit
\item Turn on the right-most zero bit in a word
\item Isolate the right-most zero bit
\item Form a mask that identifies trailing zeros
\item Absolute value function
\item Test if the number of leading zeros of two words are the same
\item Test if the number of leading zeros of a word is strictly less than of
  another work
\item Test if the number of leading zeros of a word is less than of
  another work
\item Sign Function
\item Floor of average of two integers without overflowing
\item Ceil of average of two integers without overflowing
\item Compute max of two integers
\item Turn off the right-most contiguous string of one bits
\item Determine if an integer is a power of two
\item Exchanging two fields of the same integer according to some input
\item Next higher unsigned number with same number of one bits
\item Cycling through 3 values
\item Compute parity
\item Counting number of bits
\item Round up to next highest power of two
\item Compute higher order half of product of x and y
\end{enumerate}

Reference implementation of those programs were obtained from the examples
directory of the stoke repository~\citep{stoke-code}.

\clearpage
\section{Examples of Hacker's delight optimisation}
\label{sec:hd_opt}

The first task of the Hacker's Delight corpus consists in turning off the
right-most one bit of a register.

When compiling the code in Listing~\ref{lst:base}, \texttt{llvm} generates the
code shown in Listing~\ref{lst:clang}. A typical example of an equivalent version
of the same program obtained by the data-augmentation procedure is shown in
Listing~\ref{lst:synth}. Listing~\ref{lst:opt} contains the optimal version
of this program.

Note that such optimization are already feasible using the stoke system of \citet{schkufza2013stochastic}.

\lstset{language=[x86masm]Assembler,
  frame=ltrb,
numbers=left}

\begin{figure}[h]
  \begin{subfigure}[b]{0.45\linewidth}
    \newsavebox{\baselistingbox}
    \begin{lrbox}{\baselistingbox}%
      \begin{minipage}{\textwidth}%
\begin{lstlisting}[language=c]
#include <stdint.h>

int32_t p01(int32_t x) {
  int32_t o1 = x - 1;
  return x & o1;
}
\end{lstlisting}
      \end{minipage}%
    \end{lrbox}%
    \usebox{\baselistingbox}
    \caption{\label{lst:base}Source.\\
    }
  \end{subfigure}%
  \hfill%
  \begin{subfigure}[b]{0.45\linewidth}
    \newsavebox{\clanglistingbox}
    \begin{lrbox}{\clanglistingbox}%
      \begin{minipage}{\textwidth}%
\begin{lstlisting}[language={[x86masm]Assembler}]
  pushq %rbp
  movq %rsp, %rbp
  movl %edi, -0x4(%rbp)
  movl -0x4(%rbp), %edi
  subl $0x1, %edi
  movl %edi, -0x8(%rbp)
  movl -0x4(%rbp), %edi
  andl -0x8(%rbp), %edi
  movl %edi, %eax
  popq %rbp
  retq
  nop
  nop
  nop
\end{lstlisting}
      \end{minipage}%
    \end{lrbox}%
    \usebox{\clanglistingbox}
    \caption{\label{lst:clang}Optimization starting point.\\
    }
  \end{subfigure}

  \vspace*{22pt}

  \begin{subfigure}[b]{0.45\linewidth}
    \newsavebox{\synthlistingbox}
    \begin{lrbox}{\synthlistingbox}%
      \begin{minipage}{\textwidth}%
\begin{lstlisting}[language={[x86masm]Assembler}]
  blsrl %edi, %esi
  sets %ch
  xorq %rax, %rax
  sarb $0x2, %ch
  rorw $0x1, %di
  subb $0x3, %dil
  mull %ebp
  subb %ch, %dh
  rcrb $0x1, %dil
  cmovbel %esi, %eax
  retq
\end{lstlisting}
      \end{minipage}%
    \end{lrbox}%
    \usebox{\synthlistingbox}
    \caption{\label{lst:synth}Alternative equivalent program.\\
    }
  \end{subfigure}
  \hfill%
  \begin{subfigure}[b]{0.45\linewidth}
    \newsavebox{\optlistingbox}
    \begin{lrbox}{\optlistingbox}%
      \begin{minipage}{\textwidth}%
\begin{lstlisting}[language={[x86masm]Assembler}]
  blsrl %edi, %eax
  retq
\end{lstlisting}
      \end{minipage}%
    \end{lrbox}%
    \usebox{\optlistingbox}
    \caption{\label{lst:opt}Optimal solution.\\
    }
  \end{subfigure}
\caption{\label{fig:stoke-run}Program at different stage of the optimization.}
\end{figure}

\end{document}